\documentclass[conference]{IEEEtran}
\IEEEoverridecommandlockouts
\usepackage{cite}
\usepackage{amsmath,amssymb,amsfonts}
\usepackage{algorithmic}
\usepackage{graphicx}
\usepackage{textcomp}
\usepackage{xcolor}
\usepackage[numbers,sort&compress]{natbib}
\def\BibTeX{{\rm B\kern-.05em{\sc i\kern-.025em b}\kern-.08em
    T\kern-.1667em\lower.7ex\hbox{E}\kern-.125emX}}
\begin{document}

\title{Predicting Outcomes in Long COVID Patients with Spatiotemporal Attention}

\author{\IEEEauthorblockN{Degan Hao}
\IEEEauthorblockA{\textit{Intelligent Systems Program} \\
\textit{University of Pittsburgh}\\
Pittsburgh, USA \\
deh95@pitt.edu}
\and
\IEEEauthorblockN{Mohammadreza Negahdar}
\IEEEauthorblockA{\textit{Research and Early Development} \\
\textit{Genentech, Inc.}\\
San Francisco, USA \\
negahdar.reza@gene.com}

}

\maketitle

\begin{abstract}
Long COVID is a general term of post-acute sequelae of COVID-19. Patients with long COVID can endure long-lasting symptoms including fatigue, headache, dyspnea and anosmia, etc. Identifying the cohorts with severe long-term complications in COVID-19 could benefit the treatment planning and resource arrangement. However, due to the heterogeneous phenotype  presented in long COVID patients, it is difficult to predict their outcomes from their longitudinal data. In this study, we proposed a spatiotemporal attention mechanism to weigh feature importance jointly from the temporal dimension and feature space. Considering that medical examinations can have interchangeable orders in adjacent time points, we restricted the learning of short-term dependency with a Local-LSTM and the learning of long-term dependency with the joint spatiotemporal attention. We also compared the proposed method with several state-of-the-art methods and a method in clinical practice. The methods are evaluated on a hard-to-acquire clinical dataset of patients with long COVID. Experimental results show the Local-LSTM with joint spatiotemporal attention outperformed related methods in outcome prediction. The proposed method provides a clinical tool for the severity assessment of long COVID.
\end{abstract}


\section{Introduction}
SARS-CoV-2 is a novel coronavirus previously unknown. The infection of SARS-CoV-2 can cause coronavirus disease (COVID-19) with clinical syndromes including coughing, headache, fever, etc. Among the patients infected by SARS-CoV-2, a significant number of them have sustained post-infection sequelae, which is known as long COVID. The patients with long COVID can present long-lasting COVID-19 syndromes for at least two months after the acute infection \citep{soriano2021clinical}. The long COVID symptoms such as fatigue, dyspnea, and memory problems could relapse, keep ongoing, or emerge in the following months and even years, which could endanger their lives \citep{blomberg2021long}. Although the World Health Organization (WHO) has updated the ICD-10 code to characterize long COVID, currently there are few tools for quantitative assessment of the severity of the disease \citep{pfaff2022coding}. Identifying the long COVID patients with high risk of death has great potential in providing in-time medical interventions and improving the patient outcome. 

The progression of long COVID has been predicted from several data modalities in previous studies. For example, Acute Physiology and Chronic Health Evaluation II (Apache II) is used to predict hospital mortality for patients with COVID-19 based on physiologic variables, age, and previous health conditions \citep{zou2020acute}. More recently, studies \citep{sneller2022longitudinal,pfaff2022identifying} show the prognostic power of electronic health record in predicting the outcome of patients with long COVID. The predictions in these studies are based on the patients’ statuses at the first admission, overlooking the longitudinal information in the patients’ follow-up visits. For patients with long COVID, the data collected from the first visit only describes the patients’ condition at the onset of the disease, containing limited information about disease progression. By incorporating longitudinal tabular data, the medical history could predict the disease progression. In addition to tabular data, temporal medical imaging data are also shown predictive for COVID-19 patients in the intensive care units \citep{cheng2022covid}. Effective integration of these longitudinal electronic health record could further improve the accuracy for outcome prediction. 

Deep learning has led to many promising applications for longitudinal modeling of medical data \citep{morin2021artificial}. In a study about Covid-19, CT images at different time points were first registered, segmented and then subtracted as residual values for consolidation and recovery assessment \citep{kim2021longitudinal}. A generative adversarial network (GAN) was proposed to synthesize the colorful fundus photograph to facilitate the longitudinal prediction of advanced age-related macular degeneration (AMD) \citep{ganjdanesh2022longl}. These two methods accept input with only two time points. To model medical data containing more than two time points and even in various length, recursive neural network (RNN) such as long-short-term-memory (LSTM) network \citep{lipton2015learning, wang2019long} and gated recursive unit (GRU) network \citep{choi2016doctor} have been applied. RNN is well-equipped with learning the short-term dependency (e.g., ordered/sequential patterns) in a sequence. Example applications include outcome predictions in diseases such as Alzheimer’s disease \citep{nguyen2020predicting}, AMD \citep{altay2021preclinical} etc. 

Attention enables the deep learning models to learn the correlation among multiple time points (or features) even in a long distance (i.e., long-term dependency). For example, a temporal attention mechanism was proposed to unravel the temporal importance of the history of measurements in making predictions with RNN \citep{lee2019dynamic}. This attention mechanism assigns feature importance based on previous longitudinal measurements, giving same importance for features collected at the same measurement time. In another word, features at the same time point receive the same weights. In other studies, a self-attention is integrated with the LSTM to extract both local and global representations from the feature space. Example applications include the modeling of sequential chest X-rays \citep{cheng2022covid} and longitudinal clinical data \citep{nitski2021long}. In these two applications, the self-attention mechanisms provide static feature importance, ignoring the temporal changes of feature importance. A two-level attention mechanism was proposed for mortality prediction of patients in intensive care unit where the visit-level and feature-level attention are calculated separately \citep{yang2022deepmpm}. In some chronic diseases such as long COVID, however, some syndromes that happen at the early stage could indicate worsening conditions in the future \citep{bocchino2022chest}. Customizing the weights of features jointly at feature-level and at temporal level could further leverage the prognostic information contained in the data.  

In this study, we proposed a deep learning-based, joint spatiotemporal attention mechanism to enable the outcome prediction of long COVID. The proposed attention assigns feature importance by jointly considering the time and features from a global perspective. The joint spatiotemporal attention can be plugged into many deep learning models such as LSTM and GRU. This enables a deep neural network to effectively capture both short-term and long-term dependencies within the multi-modal time series. We evaluated our method on a clinical trial dataset composed of patients hospitalized due to severe COVID-19 pneumonia. The proposed methods were also compared with state-of-the-art deep learning methods and a clinical method. Our contributions can be summarized as following:
1). We integrated multiple data modalities to predict the outcome of patients with long COVID.
2). We proposed a joint spatiotemporal attention mechanism for deep learning-based longitudinal prediction models. 
3). We evaluated the proposed method on a cohort of patients hospitalized due to long COVID.

\section{Methods}
\subsection{Problem Formulation}
Consider a sequence of longitudinal data \(\{x_{i}  | x_{1},x_{2},\cdots,x_{T}\}\), where \(T\) is the length of the sequence and \(x_{i}\) denotes the data at the \(ith\) time point. Here \(x_{i}\) can be a vector, matrix or tensor depending on the data type. The goal of longitudinal prediction is to use this data to predict a clinical outcome such as the overall survival or progression-free survival. In the context of predicting mortality of long COVID using longitudinal medical data, the problem can be formulated as a sequence-to-one problem.
\subsection{Joint Spatiotemporal Attention}
The sequence-to-one problem for long COVID mortality prediction is challenging in that there are various syndromes/features at different time points. And the syndromes/features at early time points might be correlated with those at later time points. When applying attention mechanisms to longitudinal analysis, previous works either maintain time-dependent feature importance that ignores feature diversity \citep{lee2019dynamic} or provide spatial feature importance that is static over time \citep{nitski2021long}. These assumptions may not hold as different features could have varying importance as time progresses. For example, for patients previously hospitalized due to COVID-19 pneumonia, a study shows the fibrotic-like abnormalities are common in three months but mostly will disappear after one year \citep{bocchino2022chest}. And consolidations disappeared in six months. In this scenario, the feature importance of fibrosis should not only change over time but also differ from that of consolidation. 

\begin{figure}[h]
\centerline{\includegraphics[width=\columnwidth]{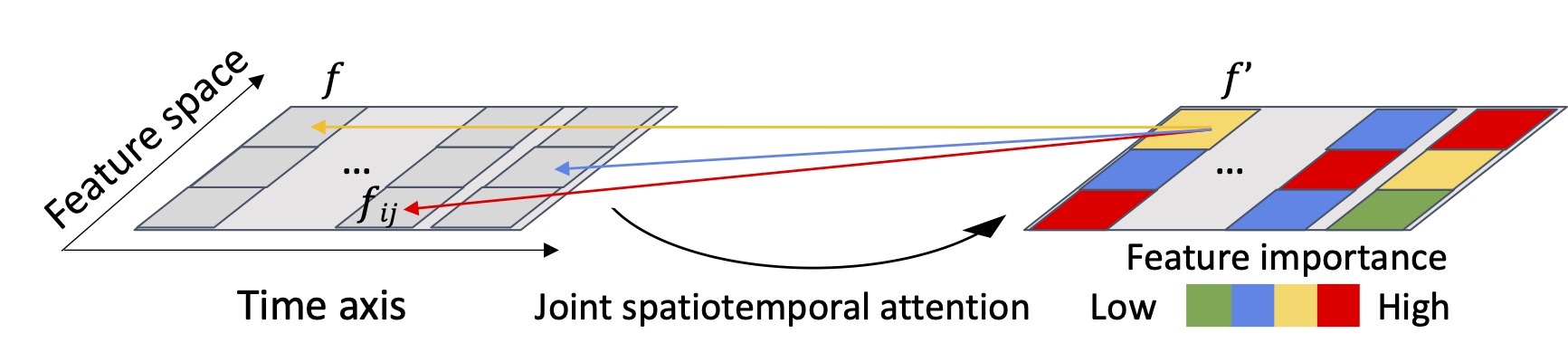}}
\caption{Illustration of the joint spatiotemporal attention. The feature importance of a feature \(f\) at a specific time point is calculated by considering each feature’s value at every time point \(f_{ij}\).}
\label{fig1}
\end{figure}

To jointly weigh feature importance over the time axis and the feature space, we proposed a spatiotemporal attention mechanism as shown in Fig.~\ref{fig1}. Specifically, when transforming the input features into key-, query- and value-space, instead of using 1-dimensional linear layers that are commonly used in machine translation \citep{sankaran2016temporal}, we adopted 2-dimensional convolutional layers that are commonly used in computer vision domain \citep{wang2018non}. The proposed attention mechanism computes keys, queries and values with 1x1 convolutional filters to calculate the alignment score as feature importance. The benefit of using 1x1 convolutional filters for attention calculation is the joint weighting of feature importance from two dimensions including the time dimension and the feature dimension. We term the convolutional self-attention that moves both along the time axis and across the feature space as the joint spatiotemporal attention. As shown in Equation 1, the hidden features are adjusted by attention with a weighting factor \(\gamma\).

\begin{equation}
f\;' = f + \gamma \cdot a(f),
\end{equation}
where \(f\) denotes the hidden features extracted by a deep neural network, \(f'\) denotes the features’ adjusted value by the joint spatiotemporal attention \(a(\cdot)\). Equation 2 shows the calculation of the joint spatiotemporal attention.
\begin{equation}
a(f) = \sum_{i=1}^{H}\sum_{j=1}^{T} softmax(q(f_{ij})\cdot k(f)) \cdot v(f),
\end{equation}

where \(H\) and \(T\) represent the size of feature space and the number of time points, respectively; the feature importance of a given feature f is determined by other features \(\{f_{ij} |  i=1,2,\cdots,H;j=1,2,\cdots,T\}\). Specifically, the feature importance is measured by calculating the alignment score via the key \(k( )\), value \(v( )\) and query \(q( )\) operations that are implemented by \(1 \times 1\) convolution.
\subsection{LSTM with Joint Spatiotemporal Attention }
\begin{figure}[h]
\centerline{\includegraphics[width=\columnwidth]{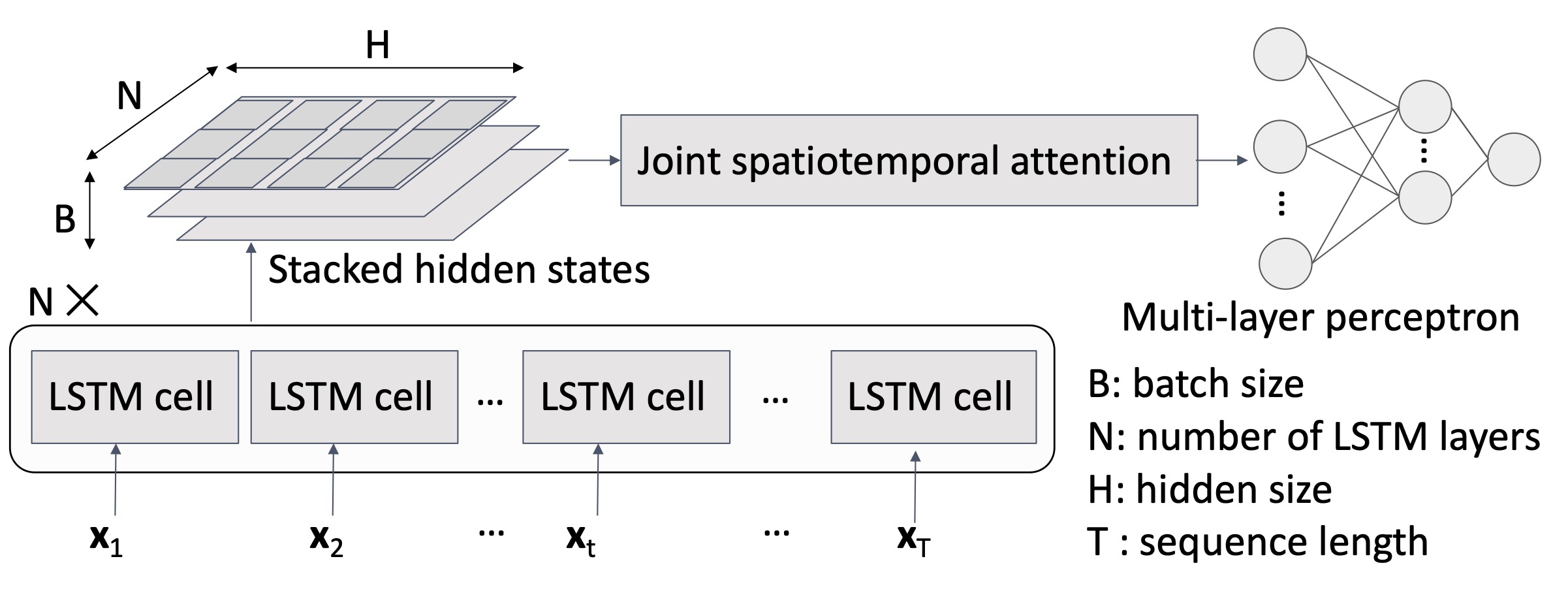}}
\caption{Model architecture of LSTM with joint spatiotemporal attention.}
\label{fig2}
\end{figure}

We first applied RNN with joint spatiotemporal attention to predict the mortality of patients with long COVID. To alleviate the vanishing gradient problem in standard RNN \citep{hochreiter1998vanishing}, we adopted the LSTM network. Fig.~\ref{fig2} shows the structure of the LSTM network with joint spatiotemporal attention. The LSTM network first extracts temporal dependency from the longitudinal input data into a feature map. The size of the feature map is \(N \times H\), where \(H\) is the number of features in the hidden state of one recurrent layer and \(N\) is the number of stacked recurrent layers. In the stacked recurrent layers, one recurrent layer takes the latent features of previous recurrent layer as input and outputs the processed signals to the next recurrent layer. In this way, the early recurrent layers extract low-level (i.e., short-term) dependency and the late recurrent layers extract high-level (i.e., long-term) dependency. We use joint spatiotemporal attention to learn the time-dependent features’ correlation. The adjusted feature map is fed to the multi-layer perceptron for outcome prediction. To evaluate the effectiveness the proposed method, we also compared it with a standalone LSTM model and a LSTM model with temporal attention \citep{lee2019dynamic}. 
\subsection{Local-LSTM with Joint Spatiotemporal Attention}
We integrated the joint spatiotemporal attention into a Local-LSTM for separate learning of short- and long-term dependency. While attention is good at learning long-term dependencies, it lacks the capability of modeling local/sequential structures in order \citep{vaswani2017attention}, which usually exist in short-term dependencies. Take the electronic health record for an example. Different from an image or a sentence where the sequential ordering of elements has contextual meanings, the electronic health record within a few days does not follow strict orders (i.e., short-term randomness). For instance, a patient could undergo the lab tests on the first day and medical imaging examination on the next day or vice versa. The randomness of orders in short term makes the attention incapable of encoding the short-term dependency in electronic health record. In contrast, LSTM can account for the local randomness of ordering because signals stored in the memory cells can still propagate even if the local order is changed. To disentangle the learning of short- and long-term dependency, we restricted the learning of short-term/ordering dependency to a set of Local-LSTMs and the learning of long-term dependency to joint spatiotemporal attention. The Local-LSTM only learns the sequential patterns within a window size and extracts the local patterns as hidden states. 

\begin{figure}[h]
\centerline{\includegraphics[width=\columnwidth]{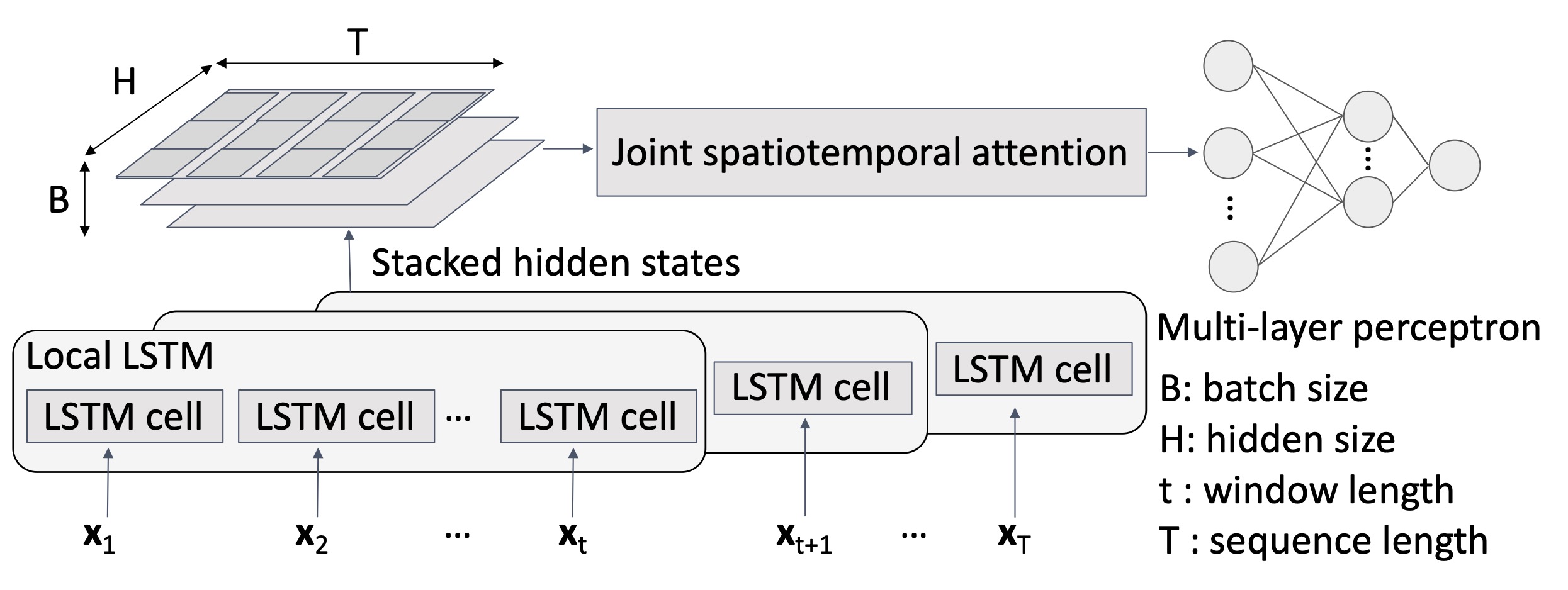}}
\caption{Model architecture of Local-LSTM with joint spatiotemporal attention. The model uses Local-LSTM to encode the short-term dependency and joint spatiotemporal attention to encode the long-term dependency.}
\label{fig3}
\end{figure}

As shown in Fig.~\ref{fig3}, the Local-LSTM sequentially processes longitudinal data in a sliding window of a given length \(t\). At each time point and its \(t - 1\) neighbors, a patient's condition is represented by a latent feature vector of size \(H\). For a sequence of length \(T\),  the shape of the sequence's latent feature map would be \(T \times H\). After concatenating the hidden states from a batch of size \(B\), the stacked hidden states become a tensor of size \(B \times T \times H\). Then the joint spatiotemporal attention refines the hidden states by mining the feature-wise and long-term dependencies and outputs adjusted hidden states. The adjusted hidden states are fed to a multi-layer perceptron for mortality prediction. Our model differs from the R-Transformer \citep{wang2019r} in that we applied joint spatiotemporal attention on Local-LSTM while the previous work used only spatial attention on Local-RNN. Our work is also different from the transformer models in computer vision \citep{dosovitskiy2020image} and natural language understanding \citep{vaswani2017attention} where the temporal information are encoded with the positional embeddings while we encode the temporal information with Local-LSTM. Due to the local randomness of the sequences, we did not experiment with the methods based on positional embeddings.

\subsection{The Clinical Model}
Apache II system is a clinical nomogram that has been widely used in clinical practice to quantify the disease severity \citep{knaus1985apache}. By measuring the initial values of 12 routine physiologic measurements, age, and previous health status, Apache II gives a score between 0 to 71 where higher scores indicate a higher risk of death. A previous study shows Apache II score is a significant prognostic biomarker of hospital mortality for patients with COVID-19 \citep{zou2020acute}. This method is initially designed for patients admitted to ICU and does not account for longitudinal changes. As a result, Apache II only models patients’ data at the admission to the hospital. We applied this model on our study cohort and used the Apache II score to predict the patients' mortality. We term this model as the clinical model.   

\section{Experiments}
\subsection{Cohort and Dataset}
Initially a total of 396 hospitalized patients with severe COVID-19 pneumonia were identified for this study. The exclusion criteria included: i) patients without chest X-ray available; ii) patients whose contacts are lost in the follow-up visits; iii) patients with sparse data. Finally, 365 patients were included for the subsequent analysis. Patient data include demographic information, medical history, chest X-ray data that are collected at their initial admission to the hospital, and longitudinal data such as laboratory test and vital signs that are collected during patients’ multiple visits to the clinics. The average number of time points for each patient is 10 with a standard deviation of 4. The patient survival statuses are collected at the 60th day after their initial admission to the hospital. More patient characteristics are available in previous work \citep{waters2023astegolimab}. 

\subsection{Data Preprocessing}
To reduce the sparse data issue, we selected prevalent variables. Specifically, we first calculated the percentage of patients having each variable. If a variable was tested by more than 95\% of the cohort, we kept it for the following analysis. The selected laboratory test variables include Fibrinogen, C Reactive Protein, Prothrombin International Normalized Ratio, Prothrombin Time, Lactate Dehydrogenase, D-Dimer, Albumin, Ferritin, Alanine Aminotransferase, Aspartate Aminotransferase, Chloride, Protein, Alkaline Phosphatase, Bilirubin, Calcium, Creatinine, Glucose, Hematocrit, Hemoglobin, Potassium, Platelets, Erythrocytes, Sodium, and Leukocytes. The selected medical history variables include hypertension, obesity, hyperlipidemia, and diabetes mellitus based on previous studies \citep{pfaff2022coding,sneller2022longitudinal}. We calculated the score of radiographic assessment of lung oedema (RALE) to characterize the disease severity from chest X-rays \citep{warren2018severity}. Measuring the degree of acute respiratory distress syndrome (ARDS), RALE has been widely used to describe the chest radiographic findings in patients positive for COVID-19. For numeric variables, min-max normalization was employed to put each variable on the same scale of zero to one. For binary variables, we represented them with zero or one. At each time point, we concatenated the longitudinal data at that time point and the static/non-longitudinal data into a vector. Forward-filling was used to impute the missing values of temporal variables.
\subsection{Implementation Details}
We implemented the proposed networks using PyTorch framework \citep{paszke2017automatic}. Stochastic gradient optimization \citep{ruder2016overview} was used for training. The dataset was split for five-fold cross validation. In light of the number of time points available, we set window size to six for Local-LSTM; for both the LSTM model and the Local-LSTM model, the size of hidden features was set to 32. With the optimized hyper-parameters, all models are trained for 50 epochs with a batch size of two. We scheduled an annealed learning rate that gradually decreased from 1e-3 to 1e-5. The model with the best performance on the validation set was evaluated on the testing set.
\subsection{Experiment Settings}
We first evaluated the prognostic values of different data modalities by incrementally incorporating them into a LSTM model. To evaluate the performance of joint spatiotemporal attention, we compared LSTM with joint spatiotemporal attention and a previous work, i.e., LSTM with temporal attention \citep{lee2019dynamic}. The  previous method has been used for survival prediction of cystic fibrosis disease. We adapted the previous method to model the mortality prediction of long COVID. To maintain the focus of spatiotemporal attention on long-term dependencies, we replaced LSTM with Local-LSTM and formed the new model named Local-LSTM with spatiotemporal attention. The performance for mortality prediction are evaluated using the area under the receiver operating curve (AUC). We reported the average AUC after five-fold cross validation.

\section{Results}
\subsection{Comparison of Different Input Combinations}
Table I shows the AUC when using LSTM model to combine differnt data modalities for mortality prediction of long COVID. As shown in Table 1, when only using the laboratory test data (in longitudinal format), the LSTM model achieves an AUC of 0.63 on the testing set. With the incorporation of vital signs (in longitudinal format), the LSTM model’s performance is increased to 0.70. These two results demonstrate the effectiveness of LSTM in modeling longitudinal data. Then we incorporated the non-longitudinal data collected at the patients’ initial admission to the hospital. The static data include demographic data, medical history data, and chest x-ray data (in the form of RALE score). The addition of static data further increases the LSTM model’s AUC to 0.73, 0.75 and 0.76, respectively. 

\begin{table}
\centering
\caption{Prediction performance of long COVID patients' mortality using LSTM model with different data modalities.}
\begin{tabular}{ p{20pt}p{20pt}p{40pt}p{30pt}p{30pt}|p{10pt}  }
 \hline
 Lab tests& Vital signs &Demographic&Medical history&Medical images&AUC\\
\hline
 X&&&&&0.63\\
 \hline
 X&X&&&&0.70\\
 \hline
 X&X&X&&&0.73\\
 \hline
 X&X&X&X&&0.75\\
 \hline
 X&X&X&X&X&0.76\\
\hline
\end{tabular}
\label{tab1}
\end{table}

\subsection{Comparison of Different Models }
Table II shows the performance of models with different network architectures for mortality prediction of long COVID. The clinical model achieves an AUC of 0.61. We use the LSTM model with a combination of all data modalities as the baseline model (AUC = 0.76). After adding temporal attention, the LSTM model’s AUC is slightly increased from 0.76 to 0.77. After replacing the temporal attention with the joint spatiotemporal attention, the LSTM model’s AUC is further increased to 0.80. By replacing LSTM with Local-LSTM, the AUC is improved to 0.87.   
\begin{center}
\begin{table}
\centering
\caption{Prediction performance of long COVID patients' mortality using different models.}
\begin{tabular}{ p{180pt}|p{30pt} }
 \hline
 Model name& AUC\\
 \hline
 Clinical model \citep{knaus1985apache}&0.61\\
   \hline
 LSTM &0.76\\
  \hline
 LSTM with temporal attention \citep{lee2019dynamic}&0.77\\
  \hline
 LSTM with joint spatiotemporal attention&0.80\\
  \hline
 Local-LSTM with joint spatiotemporal attention&0.87\\
  \hline
\end{tabular}
\label{tab1}
\end{table}
\end{center}

\section{Discussion}
In this study, we proposed a joint spatiotemporal attention mechanism to enable the simultaneous learning of feature importance across the temporal and feature space. The integration of joint spatiotemporal attention into the LSTM model and the Local-LSTM model demonstrated the effectiveness of the proposed attention mechanism. We further evaluated the LSTM model and the Local-LSTM model by comparing with related methods. The experiments on a clinical dataset demonstrated the proposed models’ effectiveness for mortality prediction of patients with long COVID.

\subsection{Effects of spatiotemporal attention}
The joint spatiotemporal attention enables dynamic feature importance both along the time axis and across the feature space. Previous work either calculates the feature importance independent of time or calculates the temporal importance being unaware of the inter-feature differences. In longitudinal modeling, the feature importance can vary both along time and by features. The proposed method addresses this problem by learning two-dimensional convolutional filters. Our experimental results show that the LSTM with joint spatiotemporal attention outperformed LSTM with temporal attention and LSTM without attention, demonstrating the effectiveness of the proposed attention mechanism. 
\subsection{Comparison of models for longitudinal prediction}
When comparing the clinical model, LSTM models with and without attention and Local-LSTM model with attention, we can see that the Local-LSTM model achieves higher performance (AUC=0.87) than the rest models. The regular LSTM model entangles the learning of the short-term/ordering and long-term dependencies together. In contrast, the Local-LSTM model enables the separate learning of short-term and long-term dependency in longitudinal data. The joint spatiotemporal attention receives latent features processed by the Local-LSTM and can only learn the remaining information about long-term dependency. Since attention is unable to model the sequential patterns in local structure, leaving the learning of short-term/ordering dependency to LocalLSTM avoided attention's drawback. This enhanced the whole model (Local-LSTM with spatiotemporal attention)'s capacity in pattern recognition of longitudinal data. The results suggest separate learning of short-term/ordering and long-term dependency could improve the outcome prediction. In addition, the clinical model does not perform as good as the LSTM model or the Local-LSTM model for many reasons. First of all, it may have to do with its linear addition of scores, which excluded nonlinear interactions among variables. Secondly, some of the variables for Apache II score’s calculation is not available in our dataset and could also limit Apache II system’s performance. In addition, the clinical model is incapable of modeling longitudinal data.

\section{Conclusion}
In conclusion, we proposed a joint spatiotemporal attention mechanism for long COVID mortality prediction from longitudinal medical data. We integrated the proposed attention mechanism to deep learning frameworks including a LSTM model and a Local-LSTM model. Our experiments show the effectiveness of the proposed method. In future work, we plan to compare our method with additional related methods and evaluate our method on additional longitudinal prediction tasks and other diseases. 

\bibliographystyle{IEEEtran}
\bibliography{conference_101719} 

\end{document}